\documentclass[sigconf]{acmart}

\AtBeginDocument{%
  \providecommand\BibTeX{{%
    \normalfont B\kern-0.5em{\scshape i\kern-0.25em b}\kern-0.8em\TeX}}}

\setcopyright{acmcopyright}
\copyrightyear{2023}
\acmYear{2023}
\acmDOI{XXXXXXX.XXXXXXX}

\acmConference[Conference acronym 'XX]{Make sure to enter the correct
  conference title from your rights confirmation emai}{June 03--05,
  2018}{Woodstock, NY}

\begin{document}

\title{Measuring Equality in Machine Learning Security Defenses: A Case Study in Speech Recognition}

\author{Luke E. Richards}
\email{lerichards@umbc.edu}
\affiliation{%
  \institution{University of Maryland, Baltimore County}
  \institution{Pacific Northwest National Laboratory}
  \country{USA}
}

\author{Edward Raff}
\email{raff.edward@bah.com}
\affiliation{%
  \institution{University of Maryland, Baltimore County}
  \institution{Booz Allen Hamilton}
  \country{USA}
}

\author{Cynthia Matuszek}
\email{cmat@umbc.edu}
\affiliation{%
  \institution{University of Maryland, Baltimore County}
  \country{USA}
}

\copyrightyear{2023} 
\acmYear{2023} 
\setcopyright{acmlicensed}\acmConference[Accepted at AISec '23]{To Appear in the Proceedings of the 16th ACM Workshop on Artificial Intelligence and Security}{November 30, 2023}{Copenhagen, Denmark}
\acmBooktitle{the upcoming Proceedings of the 16th ACM Workshop on Artificial Intelligence and Security (AISec '23), November 30, 2023, Copenhagen, Denmark}
\acmPrice{XXXX}
\acmDOI{XXXX}
\acmISBN{XXXX}

\renewcommand{\shortauthors}{Anonymous}

\begin{abstract}
Over the past decade, the machine learning security community has developed a myriad of defenses for evasion attacks. An understudied question in that community is: for whom do these defenses defend? This work considers common approaches to defending learned systems and how security defenses result in performance inequities across different sub-populations. We outline appropriate parity metrics for analysis and begin to answer this question through empirical results of the fairness implications of machine learning security methods. We find that many methods that have been proposed can cause direct harm, like false rejection and unequal benefits from robustness training. The framework we propose for measuring defense equality can be applied to robustly trained models, preprocessing-based defenses, and rejection methods. We identify a set of datasets with a user-centered application and a reasonable computational cost suitable for case studies in measuring the equality of defenses. In our case study of speech command recognition, we show how such adversarial training and augmentation have non-equal but complex protections for social subgroups across gender, accent, and age in relation to user coverage. We present a comparison of equality between two rejection-based defenses: randomized smoothing and neural rejection, finding randomized smoothing more equitable due to the sampling mechanism for minority groups. This represents the first work examining the disparity in the adversarial robustness in the speech domain and the fairness evaluation of rejection-based defenses.
\end{abstract}



\maketitle

\section{Introduction}

Systems integrating machine learning (ML) introduce a new attack surface for adversaries to take advantage of with regards to security. So far, we observe that when developing defenses for these systems, only a few works take any metric beyond the increase in aggregated adversarial robustness into account. However, these defenses are rarely considered when designing systems that interact with humans. In a field where many defenses are already not adequately evaluated, leading to a false sense of security \cite{carlini2019evaluating}, such user-centric evaluations should also be at the forefront. In this work, we expand on the measurement of machine learning security in a user-centric manner. 

We do this by examining the broader set of defenses being introduced in machine learning systems and showcasing how they can be biased and tested for such bias. These include defenses that exist outside of the model weights themselves, like preprocessing and postprocessing. We integrate concepts of parity studied by the machine learning fairness community to measure the equality of performance and outcomes resulting from such mitigation methods. Through this intersectional view, we seek to aid the understanding of the questions the community should ask when integrating various defense defenses. Broadly, this question presents itself: who do these proposed defenses work for when the system is under attack and when not? 

This comes at a time when machine learning security is being recognized and reaching the maturity of deployment in the private and public sectors \cite{grosse2023machine}. While there exist many complex threat models and levels of access in the literature, we examine impact-based attacks where a defender would attempt to protect the system from potential adversarial inputs. This type of attack represents a level of access only at deployment time rather than the development of the model. Commonly, the success criteria of an introduced defense are high performance with and without the presence of an attack. In this work, we expand this success criterion to ensure the defenses work for many different user demographics with and without attack and across the spectrum of attack strengths.     

This historical monolithic sample evaluation method ignores a recurring real-world phenomenon of inequality that such algorithmic methods may present. To address this, we first formally define two metrics per demographic group, analyzing disparate defense in robustness training and preprocessing-based and disparate false rejection in popular proposed postprocessing defenses. We then present the case for a more comprehensive evaluation of defenses to account for their downstream implications, especially as such defenses are deployed in critical safety environments.    

\begin{figure}
  \includegraphics[width=\columnwidth]{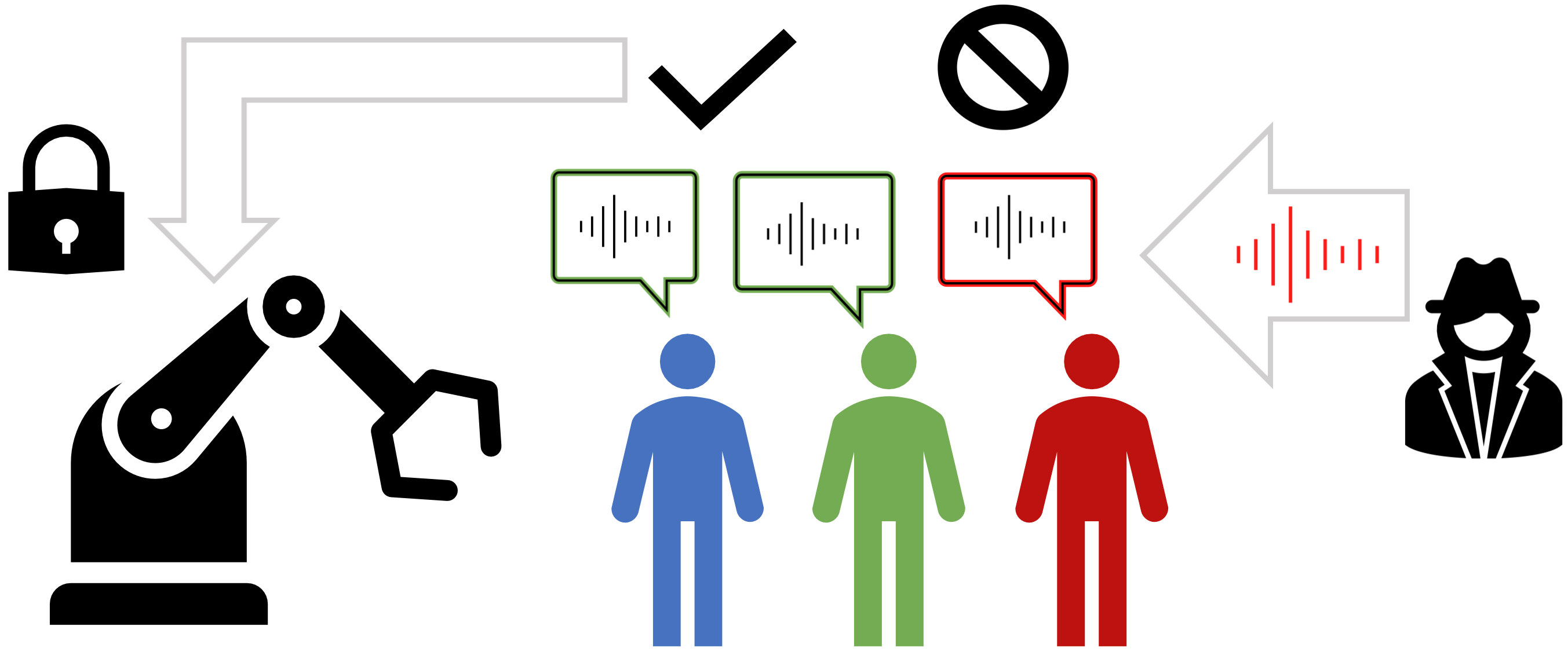}
  \caption{A depiction of machine learning models operating on a robotic arm accepting or correctly classifying speech commands from some demographics (depicted by color) but not others under attack by a theoretical adversary.}
  \vspace{-10pt}
  \label{fig:scenario}
\end{figure}

The need for such a proposed evaluation can be exemplified by imagining a manufacturing or delivery fulfillment scenario where a worker attempts to give a speech command to a robotic operation machine \cite{richards2023DEIHRI} (Fig. \ref{fig:scenario}). Due to the safety concern of the incorrect machine operation being performed, which poses risks to humans, the manufacturer added a rejection-based defense method to avoid acting on adversarial commands. The defense offered here is rejection or abstinence, a standard method introduced in defense literature \cite{cohen2019certified,melis2017deep, crecchi2022fader}. The speech command recognition model was never tested with a diverse range of end-users and only reported general accuracy for classifying attacks. As such, it systematically rejects men in their twenties with Indian accents as the model perceives them as potential adversarial examples. 

Such a scenario can illustrate the dire need for an evaluation standard, considering how rejection methods perform on various end-users and how defenses add to the impacts that automated technology can have. Notably, to our knowledge, no existing work on rejection methods has proposed or performed such an evaluation. This comes at a time when society is becoming aware that many have faced technology not recognizing them \cite{buolamwini2018gender} or understanding their speech. Here, the machine learning security methods add another layer of complexity that requires analysis that the community must integrate. There is a lack of benchmarks and taxonomy for how such defenses can also present unfairness. We propose a framework for analyzing this disparity within the taxonomy of current defense types. In our case study on speech command recognition, we showcase how such biases can be found in robustness training (adversarial training, robustness architecture tricks, and data augmentation) and rejection-based defenses (randomized smoothing and neural rejection). We outline how the exact measurements can also be applied to preprocessing, such as denoising. 

Our contributions are the following: 

\begin{itemize}
    \item Relating parity measures for equality in adversarial defenses in robustness training, rejection, and preprocessing defenses.
    \item Identifying two user-centric task-based speech command recognition datasets with social labels that avoid the bias of common fairness and robustness evaluations in gender classification through facial features. 
    \item Conducting two case studies across two speech recognition datasets:  1) on how robustness training methods impact equality of defenses and 2) a comparison of rejection-based methods and their fairness implications. Offering an empirical analysis of the complex relationships and straightforward methods for increasing parity across age, accent, and gender groups. 
\end{itemize}

\section{Related Work}

\subsection{Machine Learning Security}

 Machine learning security is a growing field \cite{biggio2018wild} due to the proliferation of machine learning models in the wild that adversaries can exploit. One of the most prototypical attacks is generating digital adversarial examples to evade a classification model. Many attack methods exist that attempt to be less computationally expensive \cite{pintor2021fast} and transfer better to other models \cite{zhang2020black}. Along with this, numerous defenses are being proposed. These include adversarial training, including adversarial examples in the training loop \cite{goodfellowAT2015}, certifying models \cite{cohen2019certified}, preprocessing \cite{zelasko2021adversarial}, and randomization \cite{guo2018countering}. While it is out of scope computationally to evaluate each of these methods, we offer a framework such that current and future defense methods can be evaluated with respect to fairness. 

\subsection{Algorithmic Fairness}

Creating models and algorithmic systems that use them to make decisions has been shown consistently to amplify and crystallize social biases. Algorithmic Fairness encompasses a body of work that addresses measuring and mitigating social biases instilled through algorithms~\cite{pessach2020algorithmic}. It would not do justice to this rich literature to attempt to summarize all works in this space, so we focus on the intersection between this body of work and security. Our work focuses on using social labels to form subsets or slices of analysis. Works such as \cite{dwork2012fairness} have pioneered this form of analysis to measure better how machine learning models fail and metrics for acceptable failure between subgroups. 

\subsection{Algorithmic Fairness and Security}
\label{section:related_work_fairness_and_sec}

Adversarial machine learning has been shown to be a beneficial tool in learning less biased representations from data \cite{zhang2018mitigating, wang2022fairness}. On the other hand, adversarial methods have been used to degrade the fairness of a model purposefully. Poisoning attacks have been introduced, which further social group disparity in equalized odds between a privileged and unprivileged group \cite{solans2020poisoning, jo2022breaking}. More complex poisoning attacks exist that focus on damaging a subpopulation's performance while preserving all others \cite{jagielski2021subpopulation}. Frameworks have been introduced to attempt to understand the effect of poisoning attacks where the adversary manipulates subgroup labels on accuracy and fairness \cite{celis2021fair}.In contrast to our work, they examined how current defenses against poisoning attacks perform. Works have also investigated subverting social attribute re-ranking in image search by modifying the image database with targeted attribute detection \cite{avijitfairimagesearch2022}. Other work has examined how similar test-time poisoning attacks can be performed on clustering algorithms \cite{chhabra2021fairness}.

However, works have only rarely examined a model's security in relation to the fairness of social factors. Prior works have examined adversarial hardening and fairness, defined as the vulnerability of each class within a classification task \cite{xu2021robust} in completely class-balanced datasets. Researchers have begun to address critical issues of imbalanced datasets \cite{wang2021imbalanced} but are still examining and remediating this for only class level. Similarly, evaluations have been proposed for measuring the robustness of data points through distance from decision boundaries and reliance on high-frequency features \cite{nayak2022_holistic}. Many have studied the trade-off when optimizing between fairness, utility, and robustness as separate concepts, not intersectional ones \cite{chang2020adversarial, moayeri2022explicit, pruksachatkun2021does}. Still, these phenomena are studied in the context of class fairness outside of social contexts. Our work begins to explore the effects on real social groups rather than proxy class labels. The metrics optimized for fairness within these works also include accuracy, which does not account for all defense methods that can have systemic rejection (as shown in Section \ref{section:rejection_framework}). There have been efforts to educate and include such intersectional analysis in the community still along the axis of fairness and robustness separated \cite{yurochkin2023ai}. 

Our work most closely compliments the work of~\cite{nanda2021fairness}, which introduces the concept of biased robustness. That is, a model has different levels of robustness for sub-populations. Within their work, they attempt to use adversarial examples as an upper bound to measure the needed permutation for an evasion attack to be successful. They defined a lower bound for robustness through a randomized smoothing certification method. While this work is a critical step for equal robustness, it audits models without a threat model present. In our work, we expand upon this concept as we acknowledge that a model creator would desire to take defense steps to protect against adversarial attacks. We are then interested in the inequality of these defenses concerning sub-population robustness. Current works have begun examining these trade-offs between fairness and robustness in tabular data in a binary fairness grouping when applying a single defense of adversarial training \cite{sun2022towards}. A handful of works examine jointly optimizing models to be fair and robust against data corruption in tabular data \cite{roh2021sample, konstantinov2022fairness}.

We note many of the works~\cite{nanda2021fairness, avijitfairimagesearch2022, jagielski2021subpopulation} in fairness and adversarial machine learning  focus on attribute classification through facial recognition. Many of these focus on gender recognition in a binary and from crowd-sourced perceived gender. While the motivation is of deep importance to make machine learning models robust for all, we suggest that gender recognition is in itself a biased task and distracts from a proper evaluation of the usability of machine learning models on a task by end-users \cite{hamidi2018gender}. We thus examine speech applications, as the social attributes used to measure fairness are not entangled with the task itself. We expand upon a dataset that can be used for the future in Section~\ref{section:dataset}. 

\subsection{Adversarial Attacks and Defenses in Speech Domain}

Our case study examines the equality of defense in the speech recognition domain. Works within this space have shown that methods used in the image domain can transfer to the speech domain in untargeted attacks \cite{gong2017crafting} and targeted phrase attacks  \cite{carlini2018audio}. Novel defenses within this space have been introduced through methods of data compression \cite{carlini2018audio}, audio denoising/purification \cite{sreeram2021perceptual}, and randomized smoothing \cite{zelasko2021adversarial}. These studies view the dataset as a homogeneous group and report solely dataset-level performance. 

\section{Evaluation Framework}

We propose an evaluation framework to analyze the bias of commonly used defenses in machine learning against adversarial interactions. We consider a model $f$ with parameters $\theta$ that takes inputs $x$ and outputs $y$. Our threat model is an adversary producing a modified input $x'$ to make the model $f$ exhibit undesired behavior. Within our framework, we analyze the worst-case scenario in which an adversary has both the model class and parameters such that they can perform a white-box attack. This gives an upper bound on the success of an attack and the potential vulnerabilities.

We extend the definition of vulnerability such that we have for some subset of $X$ in the evaluation set labels $S$ that correspond to different social sub-populations of humans. We define biased defenses as there being systemic disparity protections under attack or the rejection of clean examples from a subgroup $s \in S$.

We differ from prior work examining biased robustness in that we assume that there is an adversary attempting to cause general, unbiased attacks on our model, such that the model creator takes steps to either 1) perform robustness training such that the model \textit{should} be robust against an attack or 2) reject the data point, in which a function attempts to categorize whether the data point is an attack. Primarily due to properties of robustness training, a model may have parity of biased robustness but would not capture the cost of overall accuracy for a subgroup $s$. The introduced metrics in the following sections help capture the trade-off between accuracy, security, and fairness of security defenses. 

\subsection{Measurements of Biased Defense in Robustness Training and Preprocessing Defense Methods}

Robustness training can take many forms. This may look like data augmentation, adversarial training \cite{madry2017towards}, large-scale pretraining with datasets attempting to capture diversity \cite{fang2022data}, and a mixture of all methods. Adversarial training offers the most well-studied empirical defense to adversarial attacks \cite{madry2017towards}. This method involves including attacked examples during training with the goal of classifying attacked examples correctly, $f(x + \delta) = y$ , despite the adversaries' optimization of $f(x + \delta) \neq y$. The method for finding such a $\delta$ has varied, with many proposals for faster and more efficient methods \cite{dong2017discovering}. 

We choose to use a strong iterative attack during training, projected gradient descent (PGD) \cite{madry2017towards}. It has been shown that most methods attempt to approximate this optimization \cite{gao2022limitations}. Learning stability is critical in the accuracy-robustness trade-off, and thus choosing larger attack budgets for $\epsilon$ can have weaker general \textit{and} attacked performance. Therefore we must choose a realistic epsilon value. However, to our knowledge, this is the first work that examines how the fairness of this accuracy-robustness trade-off disparately impacts real social groups in non-tabular datasets. 

We examine the differences between the regularly trained model and models with various robustness defense defenses across the various sub-populations. We measure accuracy across sub-populations, leading to measuring the Accuracy-Parity (AP). We examine the disparity that such defense defenses introduce through ablation studies. We note this metric should reflect the performance metric for the task, such as word error rate (WER) for speech recognition or equal error rate (EER) for speaker verification. 

We measure the area under the curve (AUC) for samples of varying levels of attack budgets $[\epsilon_{min}, \epsilon_{max}]$ and the resulting accuracy for a slice of data $(X_s, Y_s)$ where $s$ is the subgroup Eq.~\ref{eq:defense_AUCACC}. This summary statistic of accuracy area under the curve accounts (${AUC}_{acc}$) for the early loss of performance for subgroups when introducing a defense. Here a lower ${AUC}_{acc}$ indicates less protection under attack. Rather than just measuring performance at attack budget, this comparison better helps capture if stronger mitigation assists with the range of adversarial attacks that could be deployed against all users. 

\begin{equation}
    {AUC}_{acc}(X, Y, s) = \int_{\epsilon_{min}}^{\epsilon_{max}}\frac{|f(X_s + \delta_{\epsilon}) = Y_s|}{|X_s|} \mathrm{d}\epsilon. \label{eq:defense_AUCACC}
\end{equation}

We are interested in understanding how these defenses cause larger gaps in the defense from defenses. To address this, we introduce the Defense Parity (DP) metric, which intuitively attempts to capture the difference in defense performance across different subgroups of potential users. We define Defense Parity as the largest difference between subgroups as we apply defense defenses. We can extend this analysis for preprocessing defenses that attempt to cleanse adversarial examples during test time. Such methods attempt to solve $f(g(x + \delta)) = y$ by introducing a method $g$, which can take many forms.

\subsection{Measurements of Biased Rejection Defense Methods}
\label{section:rejection_framework}

When accounting for the rejection of a sample through a mechanism for classifying adversarial examples, the bias can be measured using the false positivity rate (FPR). We measure the FPR for each group $s \in S$ as a clear metric for whether a single group is being wrongly flagged more often as an adversarial attack. The simplicity of this approach extends to not needing an adversary present in evaluation, thereby limiting the technical implementation bias of attack strength or method. This is particularly appealing as many flaws can be found in how current defenses are evaluated~\cite{gao2022limitations, carlini2019evaluating}.    

We define $f^*$ as a classifier that produces $\hat{y}$ the class prediction or an abstain signal ($-1$) given the probability does not reach the threshold $\alpha$. We measure the AUC of FPRs for each group $s \in S$ with corresponding data $X_s$ over threshold values $[\alpha_{min}, \alpha_{max}]$ (Eq.~\ref{eq:fprauc}). We then examine the FPR parity  between subgroups by measuring the largest difference between subgroups. These metrics capture an ideal case with a stable increase as the security threshold $\alpha$ increases for all groups $s \in S$. The goal is to decrease the ${AUC_{FPR}}$ in general while increasing the parity.     

\begin{equation}
    {AUC_{FPR}}(X, s) = \int_{\alpha_{min}}^{\alpha_{max}} \frac{|f^*(X_s, \alpha) = -1|}{|X_s||} \mathrm{d}\alpha. \label{eq:fprauc}
\end{equation}

Multiple rejection-based methods have been proposed within the adversarial machine learning literature. However, no work so far has examined this rejection mechanism in the context of social subgroup bias. For our case studies, we compare two approaches that use different methods for determining rejection. These metrics allow us to understand first the general overall performance for each subgroup and then the overall equality of defense for groups. We note that high FPR parity (FPRP) with high false rejections ${AUC_{FPR}}$ may occur and may be desired if no alternatives exist. Ideally, the algorithm should provide high FPR parity with low false rejection scores per group ${AUC_{FPR}}$. 

\section{Experiments}
\label{section:experiments}

In this section, we outline our experiments in the domains of keyword recognition speech models across two datasets, AudioMNIST and Common Voice Clips. Each dataset presents two security settings where we train on known users and deploy for the same users (AudioMNIST) and an unknown train/test population (Common Voice Clips). We cover how we harden each model through adversarial training configurations and choices. We then outline the two methods we compare for rejecting users falsely for being detected as adversarial attacks.

\subsection{Datasets}
\label{section:dataset}

Our desired qualities from a dataset are the following: 1) lightweight and offering more minor compute costs for analyzing the equality of defenses and their combinations, and 2) being a classification dataset rather than a sentence-level automatic speech recognition, which has a higher cost in computation and requires expertise to train. Classification, in particular, has been the only task studied in rejection methods. Adapting rejection methods would be an effort, and we opt to study existing defenses. The size of the datasets should allow researchers to run more extensive exhaustive empirical experiments regarding computing-intensive methods such as adversarial training and attacks. While datasets such as \cite{hazirbas2021casual} offer the criterion of socially labeled data, it is a more complex task than speech-to-text. It is currently not computationally feasible to run such factor analysis. 

We stress that the standard classification tasks on non-tabular data are typically facial trait recognition (gender, age, etc.) which has a high existing social bias (Section~\ref{section:related_work_fairness_and_sec}). We argue speech recognition is a more deployed and user-facing technology where users of such a system would feel such discrepancies in performance compared to attribute facial recognition typically being centralized in use. We acknowledge that there are tasks such as facial recognition for security access; however, the datasets are limited and demographic labels are not readily available or provided by users like those in the speech recognition datasets we propose. 

Given these desired feature factors, we identify and showcase how unequal defenses can present across two speech recognition datasets. First, we examined a smaller dataset, AudioMNIST \cite{becker2018interpreting}, a spoken number dataset of numbers between 0 to 9. The dataset comprises 60 users with self-identified ages, accents, binary native speaker statuses, and genders. Most speakers are non-native speakers with German accents, men in their twenties. For more details of the distribution, we defer to the original paper. Each user has the same number of examples contributed to the dataset. We perform a random train/validation/test split, ensuring each user has equal representation and class representation with a ratio of 7:1:2. 

To analyze the case where we do not have user guarantees and a larger dataset, we use Common Voice Clips \cite{ardila2019common}, which has single-word phrases in English (retrieved June 2022, version 9). We focus on the English subset, which has 15,115 training examples, 7,634 validation examples, and 7,640 test examples. We remove all examples of the command ``Firefox'', leaving 13 classes total (``hey'', ``yes/no'', and the numbers 0 through 9). Common Voice allows contributors to self-identify their age, gender (labeled as the typical term for sex within the dataset), and accent. The gender ratio within the train set is 25\% female, 71\% male, and 2\% non-binary. There are 14 self-identified English accents in the train/test set. The majority of the dataset is represented by those identifying as United States English accents (47\%). Age distributions are heavily weighted toward people in their twenties (42\%). 

While this lack of representation in the dataset is not ideal, this is a typical case for datasets that drive models within speech today. When examined at the intersection of gender, accent, and age, the dataset contains 87 unique groups in the training set and 49 in the test set. There are 14 unique groups not present in the training set but within the test set. This introduces an interesting problem of generalization to variations based on just the intersections that are in the training set. The average length of a file is $0.0208$ milliseconds.

\subsection{Biased Defense in Adversarial Robustness Defenses Study}

Our case study focuses on an efficient one-dimensional convolutional network modeled after the M5 architecture \cite{dai2017very}. Additionally, we evaluate a version of M5 with a handful of tricks shown to increase adversarial robustness by removing batch normalization \cite{wang2022removing} and using a different activation function SiLU \cite{hendrycks2016gaussian}. We name this version M5-Tricks due to the bag-of-tricks approach taken (inspired by \cite{lechner2023revisiting}). 

We down-sample from the original sample rate of 48 kHz to 16 kHz. We run ablations on adding data augmentation as  prior work \cite{rebuffi2021data} has found it essential for the accuracy-robustness trade-off. We add Gaussian noise with a mean of 0 and varying standard deviations of $\{0, 0.01, 0.05, 0.1\}$ for augmentations. We do not perform augmentations during validation to ensure we choose the best clean performance model. We perform ablation studies by training a model with and without adversarial training and with and without noise augmentation. For every study, we analyze the subgroups of accent, age, gender, and if they are native speakers (only for AudioMNIST). 

For adversarial training of the model, we incorporate adversarial examples crafted by a PGD attack with $\epsilon = \{0.001, 0.005, 0.01, 0.1\}$, 10 steps, and an $\alpha = \epsilon / 5$). We add adversarial examples in the same batch as our clean examples and optimize for the joint loss with $\lambda=.5$, weighting both the adversarial and clean performance equally. We perform the same evaluation on the validation set for model selection to optimize for the model's clean and attacked accuracy. 

For our Common Voice Clips studies, we analyze three classes of models, the previously described CNN, a version of the M5 model with a handful of tricks for adversarial training, and models with and without noise augmentations. For our adversarial robustness evaluation, we run attacks on all models with a range of attack budget strengths $\epsilon = \{0.0001, 0.001, 0.01, 0.1, 0.2, 0.3\}$ and 100 steps, and $\alpha = \epsilon / 5$. We verified 100 steps were sufficient for attack convergence and enabled us to maximize our compute efficiency when running multiple configurations of experiments. Since some configurations have high noise augmentation and a high adversarial budget, we remove any model from the analysis that performs random guessing on the evaluation set.

In summary, we examine how the equality of defenses is impacted by noise augmentation with varying standard deviations, adversarial training tricks, and adversarial training with budgets. Our resulting training runs leave us with 40 models for Common Voice Clips and 40 models for AudioMNIST. Please see the appendix for all models listed. 
 
\subsection{Biased Rejection in Defenses Study}

We compare two proposed methods in the literature for their biased rejection based on the measurement of false positivity rate. Unless otherwise specified, we use the M5 model architecture and training procedure outlined above without adversarial training or noise augmentation.

For our first rejection method, we analyze the work of \cite{melis2017deep} by doing neural rejection (NR). This framework assumes we have a deep neural model $f$ with layers $l_0, ..., l_N$, each taking in the previous representation $h$. It is also assumed that the final layer, $l_N$, is a fully connected linear layer, which takes a learned representation $h_{N-1}$. They propose learning a Support-Vector Machine (SVM) with an RBF kernel  using $h_{N-1}$. Due to the SVM being a Compact Abating Probability (CAP) model \cite{melis2017deep}, we can use the probabilities as a measurement from the training distribution. This allows rejecting based on some threshold $t$, assuming it is an adversarial example. We measure the equalized odds by measuring the percentage per each sub-population in $s \in S$, which are rejected at varying thresholds $\sigma$.

The second method we analyze within the framework is randomized smoothing (RS) \cite{cohen2019certified}. This method has been widely adopted as it offers certifiable robustness with a simple implementation. Randomized smoothing can be defined as adding noise $\epsilon$ to the classification $f(x + \epsilon) = y$ where $\epsilon$ is drawn from $N(0, \sigma^2I)$. The intuition is that adversarial examples at test time would be `drowned out' with this noise addition, and the model is trained with such noise. The original work offers a simple algorithm that includes a reject from classifying task based on a Monte Carlo sampling. The prediction classes for each data point with the same noise profile addition are accumulated as counts. The top two class counts are then used to parameterize a binomial test with a threshold $\alpha$.

We measure the number of examples from a  subgroup that abstained from classification for a smoothed model. Again, we measure the FPR of rejection in case no adversarial examples are present. We use both studies' areas under the curve as a summary statistic. Again we assume that a lower ${AUC}_{FPR}$ indicates a lower FPR rate for a subgroup. Our ideal case would be that given varying levels of security (thresholds), we would see equal rejection levels per subgroup (FPR parity). 

For our NR rejection methods, we use the method of \cite{melis2017deep}, with an SVM with an RBF kernel. We use the final layer representation from the models to train these rejection models that can also perform classification. For our studies on rejection with RS, we train two M5 networks with $\sigma = .1$. We choose models based on a held-out validation set augmented with the same noise level. We then use varying numbers of samples $N = \{10^1,10^2,10^3,10^4\}$ for the binomial test. This adds a parameter that neural rejection does not offer, which we use to explore the implications of fairness of defense.  For both methods, we evaluate the $\alpha$ values between $[10^{-3}, 10^{0}]$ with a step size of $10^{-3}$. This gives us a high-fidelity view of the behavior of false positive rejection across security levels through the use of thresholds. We then compare the two methods on fair rejection parity and report our findings. 

\section{Results}

We outline the results from our meta-analysis of mitigation and their correlations to disparity for robustness training in Section \ref{results_robustness_training}. We then provide a comparison of disparity induced by neural and randomized smoothing rejection-based methods. We analyze results per dataset and then finish each subsection with a joint analysis of results when comparing the different threat models of user overlap. 

\subsection{Robust Training}
\label{results_robustness_training}

We use Pearson correlation to measure how security defenses have implications for the fairness of performance across subgroups. We encode all defenses as a binary value (defense applied or not) and report how these methods correlate with higher defense parity per each subgroup. We then analyze how varying levels of defense relate to defense parity. We find a complex relationship across datasets dependent on the assumption of user coverage and dataset complexity. 

\subsubsection{AudioMNIST}

For our analysis, we found that 15 of the 40 (about 37\% of the models trained) had a clean performance at random guessing (10\% accuracy). These models resulted from high adversarial training budgets above $0.005$ with various noise profiles. Since a model that makes random guesses can withstand any amount of noise, we removed such models from the rest of this analysis.   

\begin{figure}[h]
  \includegraphics[width=\columnwidth]{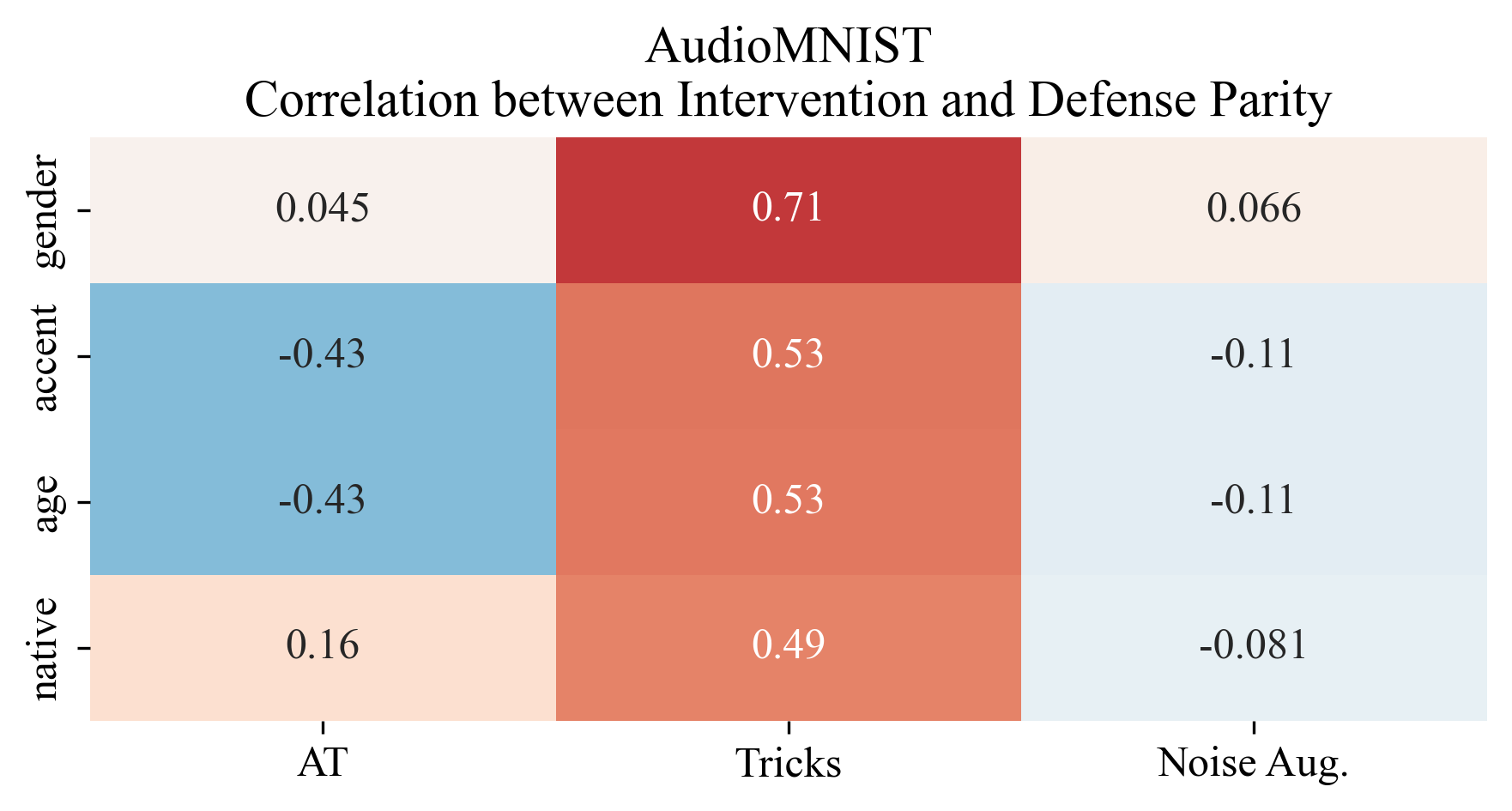}
  \caption{Pearson correlation coefficient between the presence of adversarial training (AT), AT bag-of-tricks (Tricks), and noise augmentation to defense parity per group for AudioMNIST.}
  \label{fig:audiomnist_defense_parity_corrs}
\end{figure}

On the smaller dataset, AudioMNIST, which is balanced with user contribution to training and test sets, we see that adversarial training has higher correlations to parity for adversarial training for accent and age groups (Figure~\ref{fig:audiomnist_defense_parity_corrs}). When applying the batch normalization removal (tricks), we notice a positive correlation indicating that removing the batch normalization mechanism increases disparity across all groups. This has intuitive reasons since learning the distribution of a closed-user dataset may be a viable method for robustness compared to the performance of the same defense in the open-user dataset (Common Voice Clips).  Such results indicate holes in the current evaluation of tricks in non-human user settings. There may be more nuance in removing this feature resulting in other better-defended models. The tricks have an average clean accuracy of $0.937 \pm 0.042$ while the non-trick version had $0.601 \pm 0.370$, accomplishing the task. However, with this increase in performance, we see more considerable performance disparities trading off utility and robustness with fairness. 

\begin{figure}[h]
  \includegraphics[width=\columnwidth]{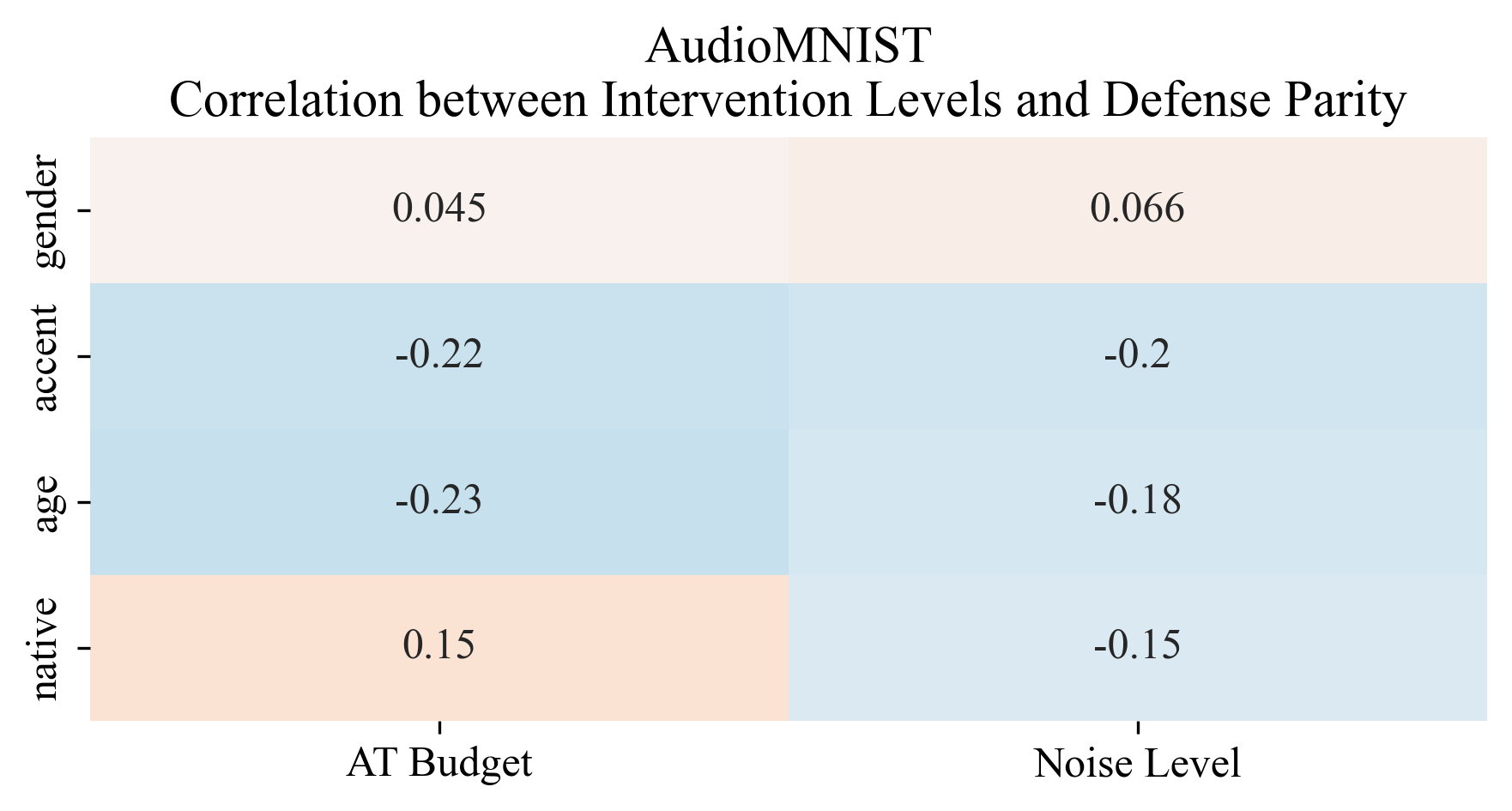}
  \caption{Pearson correlation coefficient between adversarial training budget and noise levels defenses and their resulting defense parity for AudioMNIST. A negative value indicates a stronger correlation with defense parity, and a positive value indicates a stronger relationship to defense disparity.}
  \label{fig:audiomnist_defense_parity_levels_corrs}
\end{figure}

Looking more closely at the level of disparity with relation to the increase of noise or adversarial budget, we can see a similar story to including either mitigation at all (Figure~\ref{fig:audiomnist_defense_parity_levels_corrs}). For noise augmentation level, we find there to be a weak or near zero (for gender groups) correlation between noise level augmentation, indicating that increasing the noise profile does not directly result in more robust models. This is again for all but the gender group where the correlation value is small. We see that the presence of adversarial training for accent and age groups was more telling of defense parity than the budget amount. We notice a non-linear relationship with the averaged performance metrics for both. This insight is another reminder of the importance of hyperparameter tuning for more robust models. 

\subsubsection{Common Voice Clips}

In our 40 trained models, only 6 performed at random guessing and were thus removed from the analysis. We found that all models trained with our maximum budget amount had the performance subdued for this dataset. Since our perturbation strategy happens pre-normalization to simulate real-world cases better, this may lead us to believe the overall sound profiles differ amoung datasets. Thus allowing larger budgets to have successful training.

\begin{figure}[h]
  \includegraphics[width=\columnwidth]{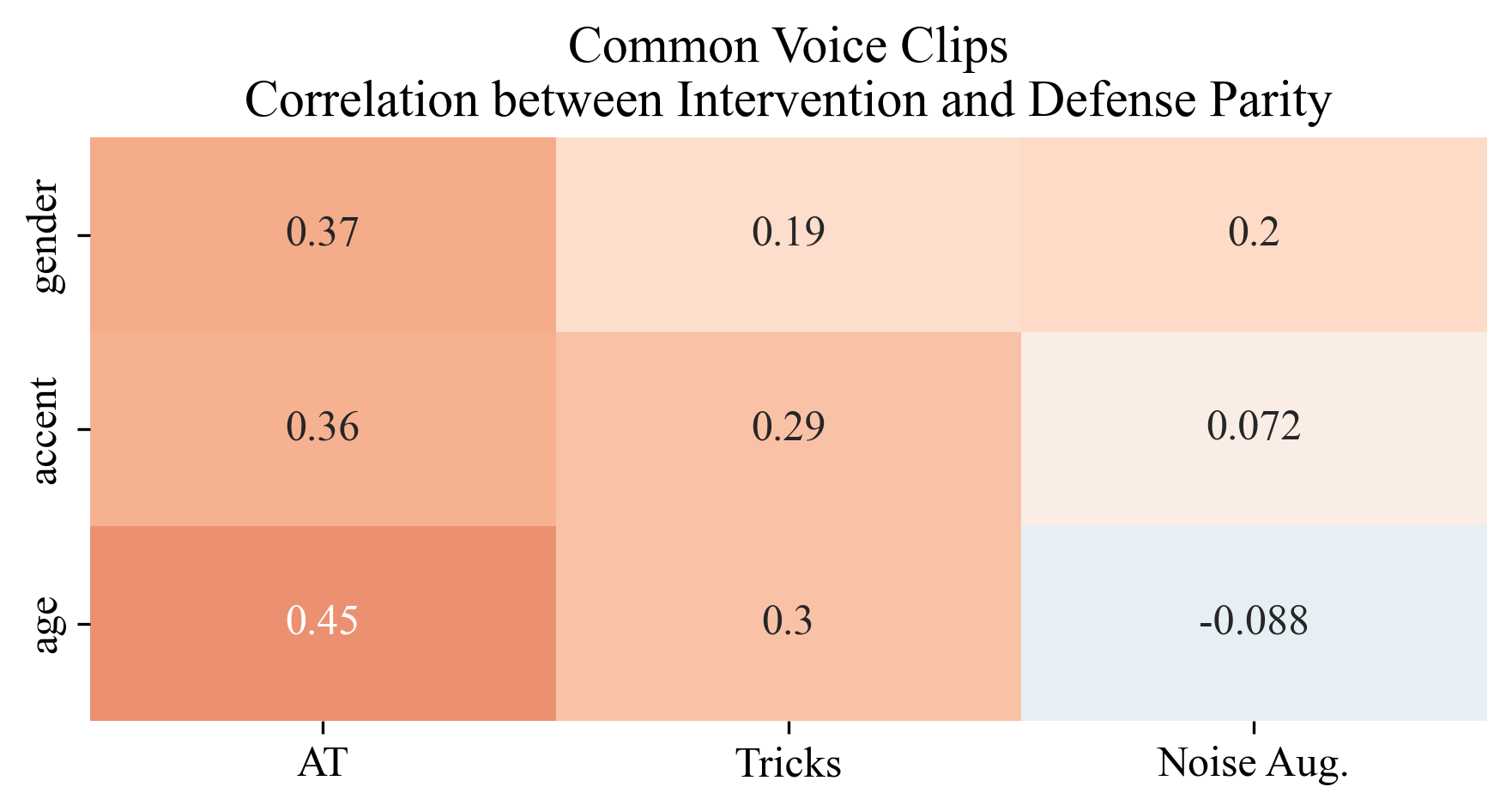}
  \caption{Pearson correlation coefficient between the presence of adversarial training (AT), AT bag-of-tricks (Tricks), and noise augmentation to defense parity per group for Common Voice Clips. A negative value indicates a stronger correlation with defense parity, and a positive value indicates a stronger relationship to defense disparity.}
  \label{fig:defense_parity_corrs}
  \vspace{-10pt}
\end{figure}

We see that adversarial training is correlated with a decrease in defense parity between all groups with a larger value for age groups (Figure~\ref{fig:defense_parity_corrs}). This contrasts with the success we saw for the application in the closed-user set. This may indicate adversarial training may only benefit all subgroups when their individual distributions do not shift between training and deployment. 

Keeping to the ability of the models to model the distribution explicitly, we again see that the removal of batch normalization (tricks) has an across-the-board correlation with defense disparity. This holding constant across these two unique datasets showcases the importance of batch normalization for fair robustness. We do not observe the trend that the model's aggregated average clean performance increases when batch normalization is removed with an accuracy of $0.654\pm 0.284$ for batch normalization and $0.633 \pm 0.287$ for models without. 

\begin{figure}[h]
  \includegraphics[width=\columnwidth]{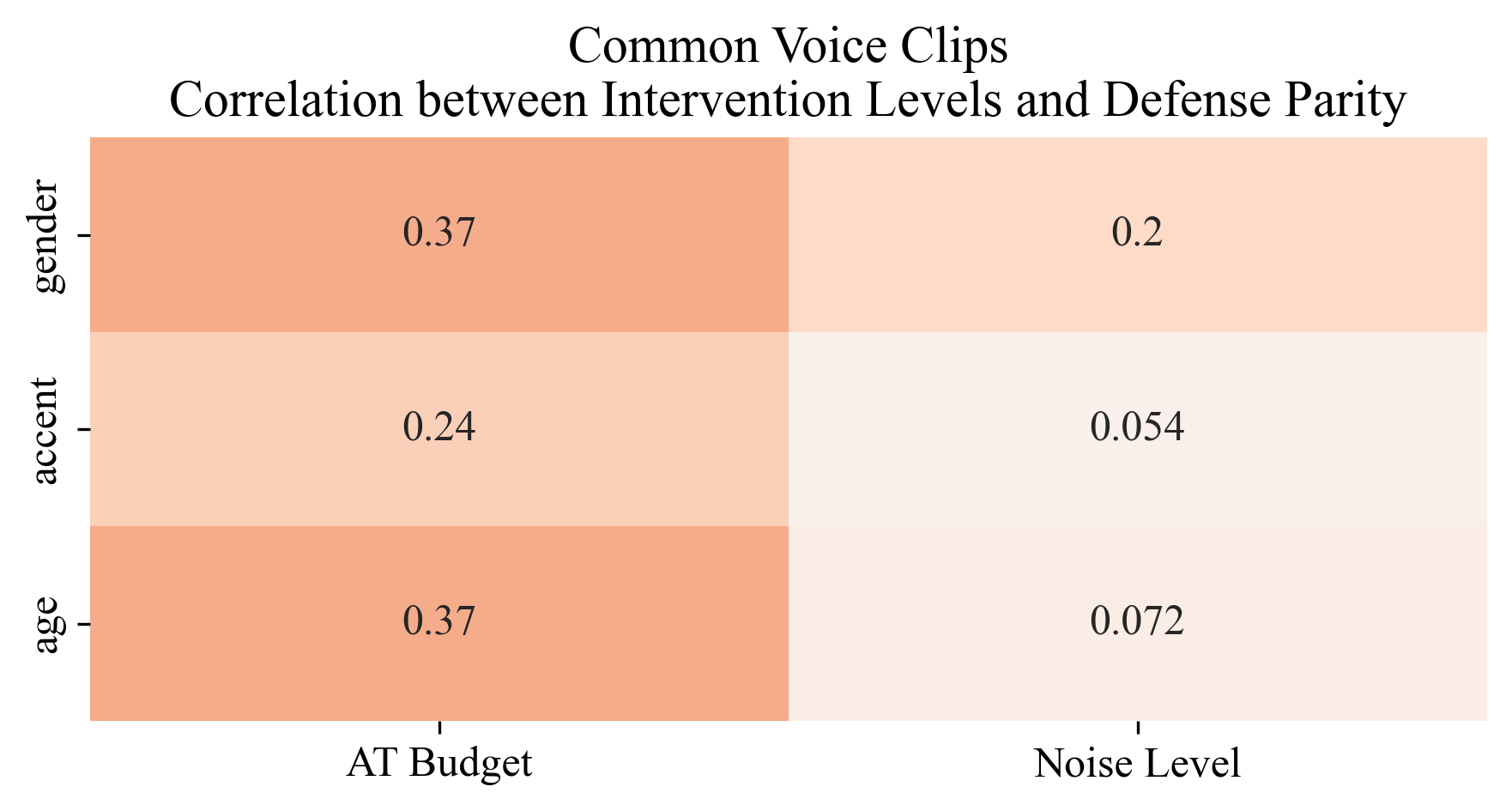}
  \caption{Pearson correlation coefficient between adversarial training budget size and noise levels augmentation and their resulting defense parity per group for Common Voice Clips. A negative value indicates a stronger correlation with defense parity, and a positive value indicates a stronger relationship to defense disparity.}
  \label{fig:defense_parity_levels_corrs}
\end{figure}

The presence of noise augmentation in training has a weaker correlation than those found with AudioMNSIT. However, for gender, we see that there is a weak correlation with disparity when having noise augmentation. When expanding this analysis beyond the binary and looking at the levels of noise augmentation (Figure~\ref{fig:defense_parity_levels_corrs}), we see this weaker correlation seen in gender holds. At the same time, accent and age groups show no correlation to the level. This, again, maybe a factor of the beneficial level being a non-linear relationship to defense parity. When examining the role that increasing the adversarial budget has on defense parity, we see that for all groups, there exists a weak correlation for defense disparity as we increase the amount of adversarial noise. This result, contrasted with AudioMNIST, shows that adversarial training tends to increase the disparity in users in the case of a larger non-overlap guarantee in users. 

\subsubsection{Joint Insights}

Our analysis of two different user scenarios has gleaned some insights into how we can design and study better systems. Given a practitioner, who is developing a dataset for a small subset of known users, such as the case of training models per user, we can see that adversarial training benefits defense parity. Noise augmentation across experiments did not have strong correlations with increasing or decreasing disparity. However, noise augmentation generally increases the performance of models under attack. Thus noise augmentation offers a robustness path that produces fair robustness. 

\subsection{Rejection: Randomized Smoothing vs. Neural Rejection}

We report the results for neural rejection (NR) and randomized smoothing (RS) using our parity measurement of the false positive rejection rate across groups over the defense level (${AUC}_{FPR}$). We find that false rejection rate disparity follows a common order across tasks of accent, age, and gender in decreasing order. Across both datasets, we can increase parity by increasing sampling in RS with no post-hoc equivalent with NR, suggesting that RS may offer more equatable rejection-based rejection. 

\subsubsection{AudioMNIST}

\begin{figure}[h]
  \includegraphics[width=\columnwidth]{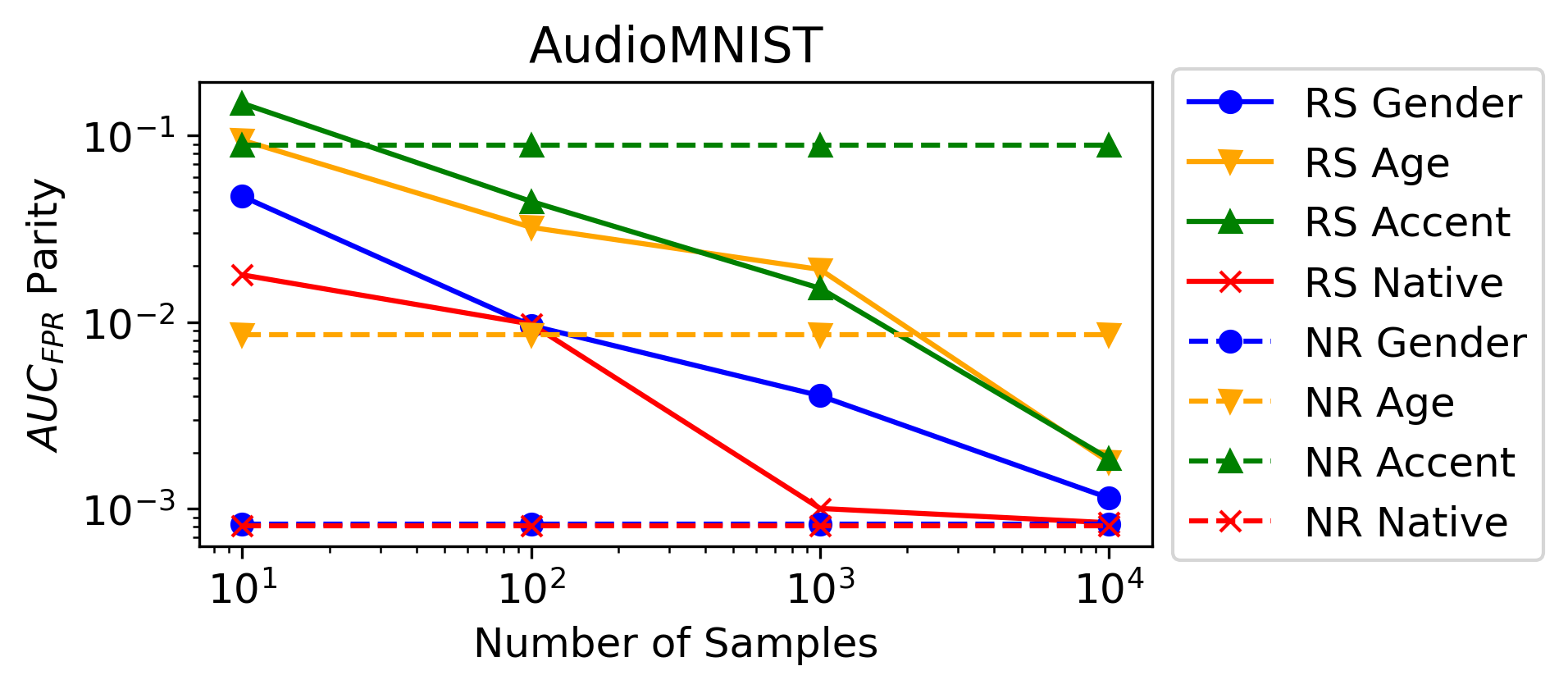}
  \caption{Parity of the area under the curve for the false positive rate ${AUC}_{FPR}$ for each method on AudioMNIST. Randomized smoothing (RS) plotted as solid lines over samples and neural rejection (NR).}
  \label{fig:audiomnist_rejection_method_comparison}
\end{figure}

We first compare the general difference between the methods in general ${AUC}_{FPR}$. NR has a value of 0.005, and RS has a value ranging from 0.074 to 0.001, dependent on the number of samples. As hypothesized, the RS sampling gives a tunable parameter for ${AUC}_{acc}$ that NR does not provide, allowing for much lower ${AUC}_{FPR}$. We contrast the parity and their correlating factors for gender, age, and accent. Notably for AudioMNIST, each user has already been seen during training. We would then hypothesize that fewer users would be falsely rejected in the case of NR due to the distribution modeling property. Ever so, this is not the case even with the underlying training distribution measurement; we see the high disparity between the accent groups (Figure~\ref{fig:audiomnist_rejection_method_comparison}). This result, along with those in the next section, indicates a more complex relationship than user representation in the case of accent and age with distributional measurement methods in speech for rejection.  

\subsubsection{Common Voice Clips}

\begin{figure}[h]
  \includegraphics[width=\columnwidth]{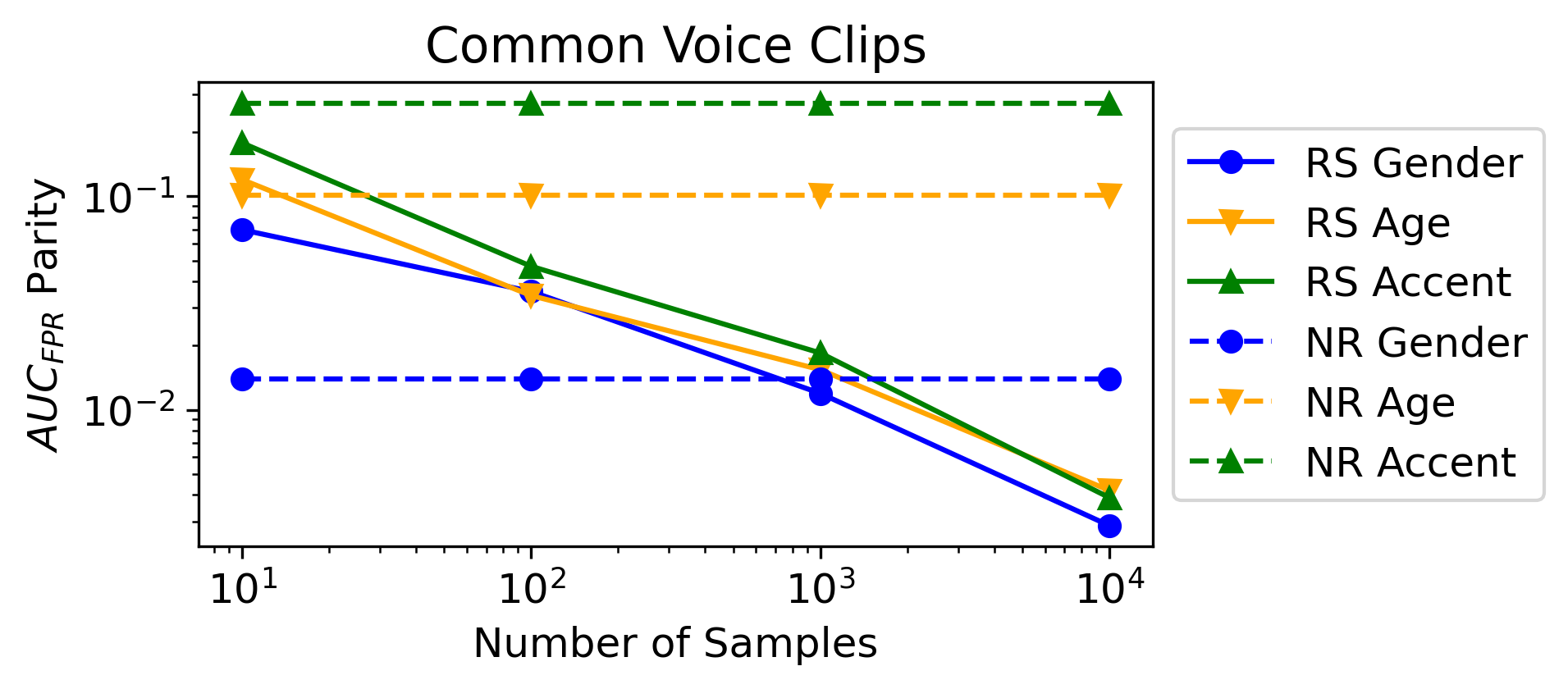}
  \caption{Parity of the area under the curve for the false positive rate ${AUC}_{FPR}$ for each method on Common Voice Clips. Randomized smoothing (RS) plotted as solid lines over samples and neural rejection (NR).}
  \label{fig:rejection_method_comparison}
\end{figure}

Our results with the NR method result in a learned model with a ${AUC}_{FPR}$ of 0.337. The RS model depends on the number of samples with values ranging from 0.116 to 0.001. This again holds that we can achieve equal or better false positive rate parity by increasing the sampling number. We then compare the false rejection parity across subgroups (Figure~\ref{fig:rejection_method_comparison}). 

For the NR method, we have ${AUC}_{FPR}$ parity values of 0.014 for gender, 0.101 for age, and a larger value of 0.273 for accent. For the RS, we see values that change oversampling values ranging from 0.069 to 0.002 for gender, 0.119 to 0.004 for age, and 0.178 to 0.003 for accent. Here we again see the benefit of choosing RS when considering the fairness implications of security methods. The benefit is that we can increase the ${AUC}_{FPR}$ parity by increasing the number of samples. When sampling $10^3$ samples, we achieved higher parity than NR for all subgroup types. While this adds to the number of model runs, this can be run in parallel and aggregated later. Future work will be needed to analyze the computational cost and latency that this may introduce at the cost of fairness in security. 

While sampling seems to increase the parity, we seek to understand better the underlying distributions for the subgroups in the NR method and factors that may contribute to the higher parity. Remember that NR relies on the fact that the SVM trained on top of the neural network will be a CAP model. Prior results from the closed-user dataset AudioMNIST indicated this to have less of a factor for user demographic groups with NR. Due to this, we investigate if having a higher subgroup representation in the training set results in higher ${AUC}_{FPR}$. Here we seek to measure the correlation between the size of subgroup training data $|X_{train, s}|$ where $s$ is the subgroup. 

We find a correlation coefficient with ${AUC}_{FPR}$ for each subgroup with the NR method. We find gender has a correlation coefficient of 0.691, age has -0.318, and accent has -0.044 for the NR method. Examining the gender groups, there is high parity in ${AUC}_{FPR}$ where a strong correlation results from only having 3 data points. Again, the accent parity proves to be the less obvious or linear relationship with the data representation. 

While the intuition data representation in RS is not as straightforward as with NR, we find that RS has more consistent correlations to subgroup training data size. For RS, the correlation with training data size and ${AUC}_{FPR}$, we see a correlation of $-0.499\pm0.08$ (averaged over all sampling). For each subgroup, we see $-0.811\pm0.16$ for gender, $-0.430\pm0.144$ for age, and $-0.652\pm0.216$ for accent. Here we see much stronger trends for having smaller ${AUC}_{FPR}$ values with increasing training data representation for subgroups. 

\subsubsection{Joint Insights}

False positive rejection poses a major emerging issue for who gets to interact with machine learning models. We showcase here across both datasets that subgroups in social groups face different levels of false rejection. Accent groups, in all cases, were the most difficult to remediate. However, randomized smoothing's sampling hyperparameter in all cases allowed us to decrease the disparity outside of also decreasing false positives for all users. A practitioner should thus incorporate rejection methods that incorporate methods with sampling as they decrease unfair, false-positive rejections. 

\subsection{Discussion}

Our framework and case study allows us to return to the original question: for whom do machine learning security methods work? We found a complex relationship between when and how adversarial training can be beneficial along with noise augmentation level for defense parity. When we had a closed-user pool, we saw a benefit with adversarial training boosting all groups and having high defense parity. When we scaled this analysis to a larger, more complex, open-user pool, our findings pointed toward noise augmentation rather than adversarial training for performance and defense parity. Future work will need to explore the role of how large-scale pretraining can provide fair robustness decreasing the defense disparity. Rather than targeting robustness, future work may also target fair adversarial training attempting to break the commonly accepted utility, fairness, and robustness trade-off working theory. 

We showcase that introducing a rejection method without evaluating in this context would result in a system that refuses to recognize speech from specific demographics more than others despite equal training data coverage. We find that rejection is based on distributional measurement methods that reject users with various accents differently and ages more than gender differences. We show through the parity measurements outlined we can alleviate this by using a smoothing method like randomized smoothing's  rejection mechanism by increasing the number of samples. Future work here will incorporate the measurement of fairness with the model's ability to reject adversarial examples. We hope such post-hoc defenses commonly deployed will incorporate fairness evaluation even with an empirically fair base model.   

\section{Conclusion and Limitations}

This work considers security defenses' fairness implications in machine learning systems. We outline the two parity metrics to help capture these implications. ${AUC}_{acc}$ (accuracy over attack budgets) measures the parity between subgroups when introducing a robustness training defense and preprocessing-based defense. ${AUC}_{FPR}$ (false rejection over levels of security) measures biases in rejection methods that do not attempt to classify adversarial attacks. We showcase how these measurements and evaluations can result in actionable studies of defenses.

We acknowledge that such a study relies on the availability of labels for subgroups that can be difficult to obtain in all domains. This limitation needs multiple efforts in collecting such labels in a scalable way and furthering proxy labeling methods. We hope this proposed method of robustness evaluation will enable such studies to be performed in other domains. Future work must integrate user-centric analysis of equality to guarantee safe, secure, and effective systems for all. This can be done by further investigating the relationship between user representations and demographic representations to performance and robustness.

\bibliographystyle{ACM-Reference-Format}
\bibliography{main}

\newpage
\appendix

\section{Appendix}

\subsection{Model Training}

Every model is trained for 1000 epochs with an Adam optimizer with an initial learning rate 1e-3. We select the model with the lowest loss on the validation set depending on the training regime. 

\section{Model Configurations Evaluated}

\begin{figure}[!h]
  \includegraphics[width=\columnwidth]{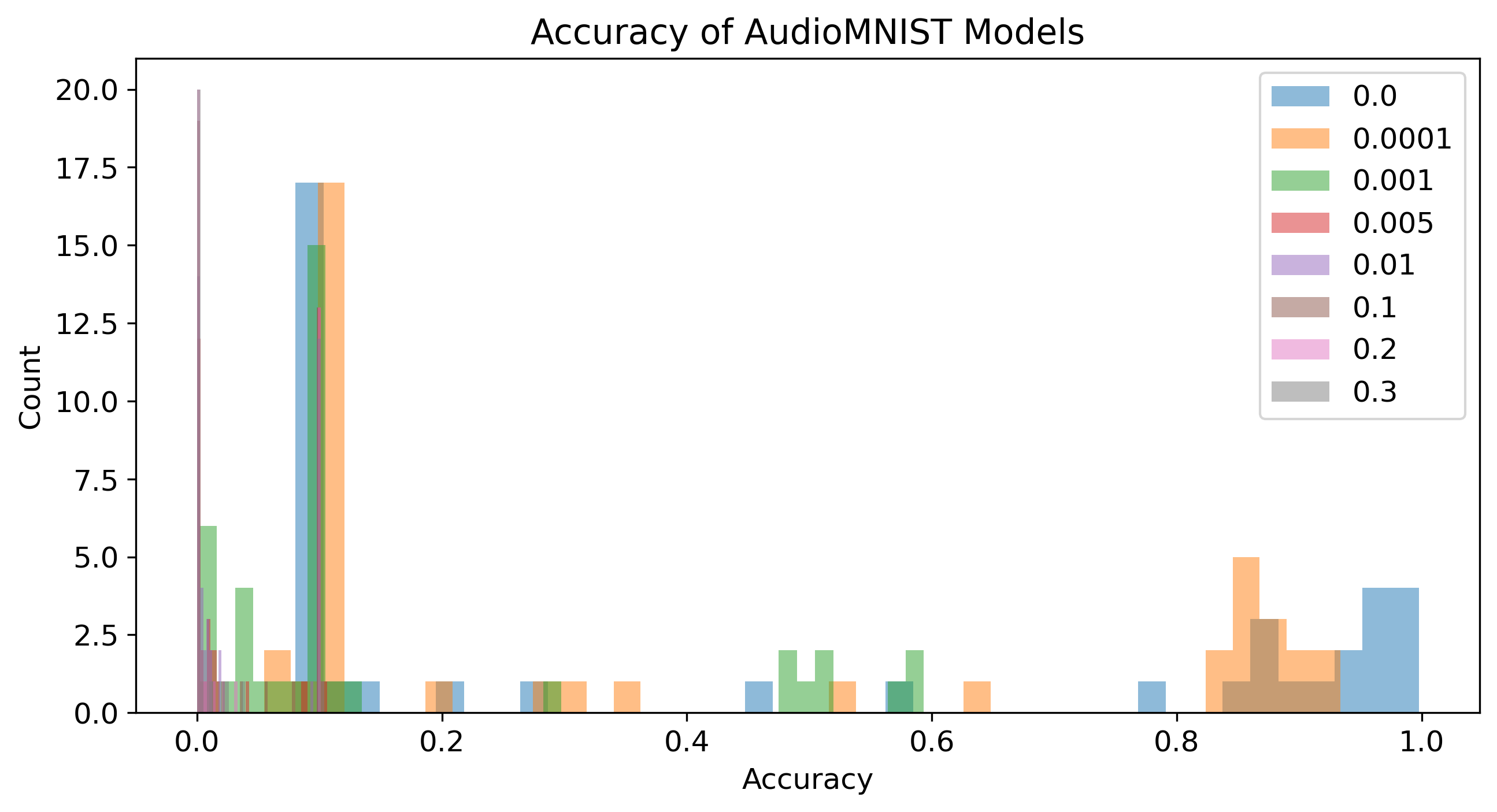}
  \caption{Distribution of model performance as a histogram for each budget of attack for AudioMNIST.}
  \label{figure:audiomnist_cvclips}
\end{figure}

\begin{figure}[!h]
  \includegraphics[width=\columnwidth]{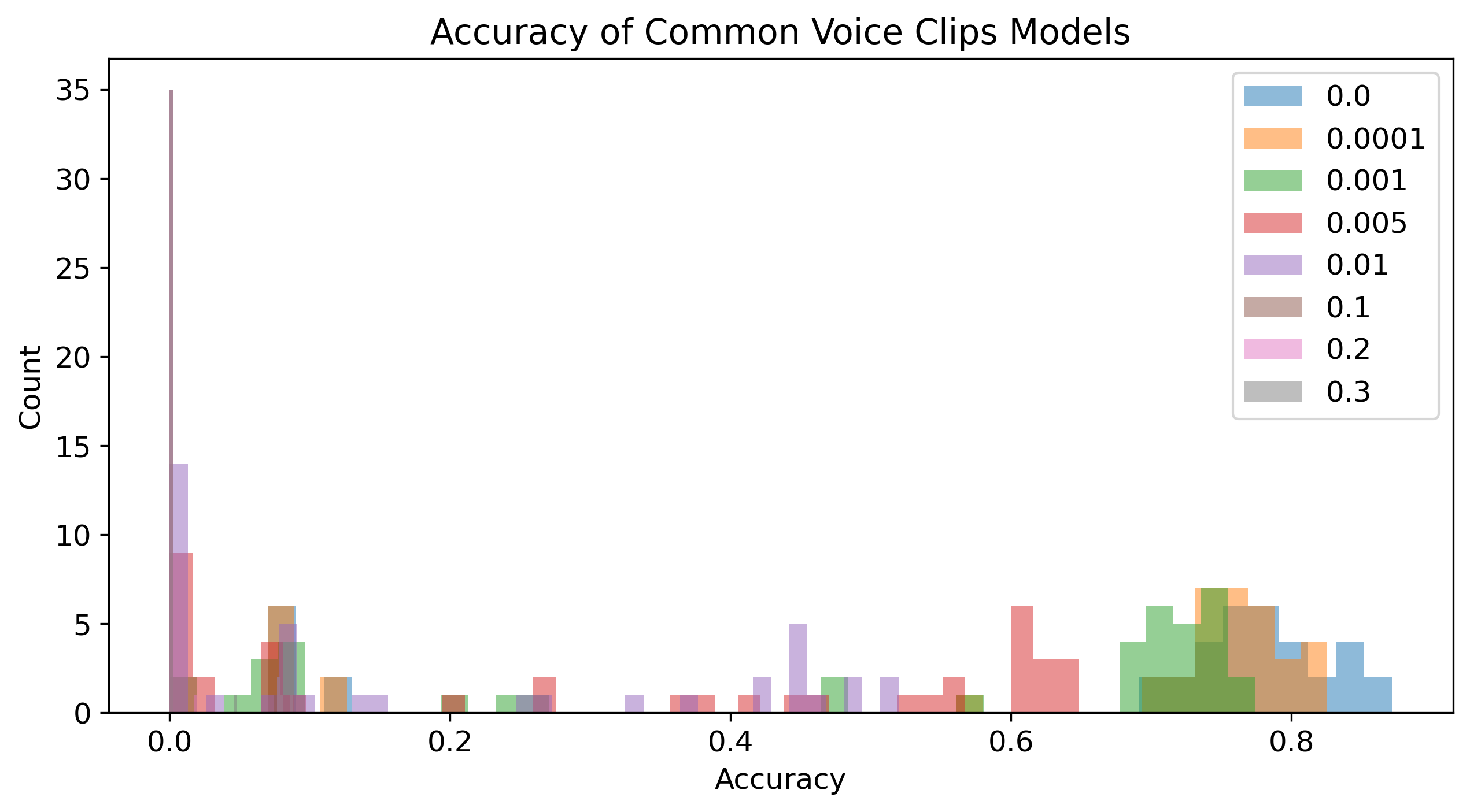}
  \caption{Distribution of model performance as a histogram for each budget of attack for Common Voice Clips.}
    \label{figure:cvclips_hist}
\end{figure}

We list all models trained for the AudioMNIST (Table \ref{table:audiomnistmodels}) and Common Voice Clips (Table \ref{table:cvclipsmodels}) and their performance for both aggregated across all groups clean accuracy and  ${AUC}_{acc}$. We as well plot histograms of the general model performance across attack budgets for visualization of performance space for both AudioMNIST (Figure \ref{figure:audiomnist_cvclips}) and Common Voice Clips (\ref{figure:cvclips_hist}).

\newpage

\begin{table}[!h]
{\caption{All models trained for the AudioMNIST dataset. Ordered by average aggregated clean accuracy.}\label{table:audiomnistmodels}}
\begin{center}
\begin{tabular}{lllll}
\toprule
Model & AT Budget & Noise Aug. & Clean Acc. &  ${AUC}_{acc}$ \\
\midrule
   M5 &    0.1 &         0.1 &  0.080333 &  0.000227 \\
 M5-T &   0.01 &         0.1 &       0.1 &      0.03 \\
 M5-T &    0.1 &           0 &       0.1 &      0.03 \\
 M5-T &    0.1 &        0.01 &       0.1 &      0.03 \\
   M5 &    0.1 &           0 &       0.1 &  0.000302 \\
 M5-T &  0.005 &        0.01 &       0.1 &      0.03 \\
 M5-T &  0.005 &        0.05 &       0.1 &      0.03 \\
 M5-T &  0.005 &           0 &       0.1 &      0.03 \\
   M5 &    0.1 &        0.05 &       0.1 &  0.006621 \\
 M5-T &   0.01 &        0.05 &       0.1 &      0.03 \\
 M5-T &   0.01 &        0.01 &       0.1 &      0.03 \\
 M5-T &   0.01 &           0 &       0.1 &      0.03 \\
 M5-T &    0.1 &         0.1 &       0.1 &      0.03 \\
 M5-T &    0.1 &        0.05 &       0.1 &      0.03 \\
 M5-T &  0.005 &         0.1 &       0.1 &      0.03 \\
   M5 &    0.1 &        0.01 &     0.101 &  0.000081 \\
   M5 &   0.01 &         0.1 &  0.102167 &  0.010445 \\
   M5 &  0.005 &         0.1 &  0.117667 &  0.001093 \\
   M5 &   0.01 &        0.05 &     0.129 &  0.011811 \\
   M5 &   0.01 &        0.01 &  0.198833 &  0.001382 \\
   M5 &  0.005 &        0.05 &    0.2845 &  0.000866 \\
 M5-T &  0.001 &        0.01 &  0.916667 &  0.004304 \\
 M5-T &  0.001 &           0 &  0.936167 &  0.006288 \\
   M5 &  0.001 &           0 &  0.946667 &  0.001786 \\
   M5 &      0 &        0.05 &     0.967 &  0.000517 \\
 M5-T &      0 &        0.05 &  0.969167 &  0.003129 \\
 M5-T &      0 &        0.01 &    0.9715 &  0.008602 \\
 M5-T &      0 &           0 &  0.973833 &  0.023675 \\
 M5-T &      0 &         0.1 &  0.975667 &  0.001859 \\
   M5 &      0 &         0.1 &  0.983167 &  0.000477 \\
   M5 &      0 &        0.01 &    0.9855 &  0.000509 \\
   M5 &      0 &           0 &  0.997667 &  0.000901 \\
\bottomrule
\end{tabular}

\newpage

\end{center}
\end{table}

\begin{table}[h]
{\caption{All models trained for the Common Voice Clips dataset. Ordered by average aggregated clean accuracy.}\label{table:cvclipsmodels}}
\begin{center}
\begin{tabular}{lllll}
\toprule
Model & AT Budget & Noise Aug. & Clean Acc. &       ${AUC}_{acc}$ \\
\midrule
   M5 &    0.1 &        0.01 &  0.069895 &  0.000327 \\
 M5-T &    0.1 &        0.01 &  0.077094 &  0.023128 \\
 M5-T &    0.1 &        0.05 &  0.078665 &  0.023599 \\
 M5-T &    0.1 &         0.1 &  0.080366 &   0.02411 \\
   M5 &    0.1 &         0.1 &   0.08233 &  0.011455 \\
 M5-T &    0.1 &           0 &  0.087696 &  0.026309 \\
   M5 &    0.1 &        0.05 &  0.124476 &  0.000315 \\
   M5 &    0.1 &           0 &  0.128796 &  0.000181 \\
   M5 &   0.01 &         0.1 &  0.710209 &  0.029066 \\
   M5 &   0.01 &        0.05 &  0.710602 &  0.028159 \\
   M5 &   0.01 &           0 &  0.722382 &  0.028335 \\
 M5-T &   0.01 &         0.1 &  0.730759 &  0.026885 \\
 M5-T &  0.005 &        0.05 &  0.743979 &  0.020492 \\
 M5-T &   0.01 &        0.05 &  0.744895 &  0.026519 \\
 M5-T &  0.005 &         0.1 &   0.74555 &  0.021992 \\
   M5 &  0.005 &         0.1 &  0.747251 &   0.02601 \\
   M5 &  0.005 &        0.05 &  0.754188 &  0.026255 \\
 M5-T &  0.001 &        0.05 &  0.756675 &  0.008198 \\
   M5 &   0.01 &        0.01 &  0.760864 &   0.02984 \\
 M5-T &      0 &         0.1 &  0.765052 &   0.00164 \\
 M5-T &  0.005 &           0 &  0.765838 &  0.016547 \\
 M5-T &   0.01 &        0.01 &   0.77055 &  0.026193 \\
 M5-T &   0.01 &           0 &   0.77199 &  0.026416 \\
 M5-T &  0.001 &         0.1 &  0.772906 &  0.005284 \\
 M5-T &      0 &        0.05 &     0.775 &  0.001862 \\
 M5-T &      0 &           0 &  0.777618 &  0.001064 \\
 M5-T &  0.005 &        0.01 &  0.785209 &  0.017105 \\
   M5 &  0.001 &         0.1 &  0.788613 &  0.011012 \\
   M5 &  0.005 &           0 &  0.796204 &  0.025012 \\
 M5-T &  0.001 &           0 &   0.79699 &  0.003247 \\
   M5 &  0.005 &        0.01 &  0.801963 &  0.025564 \\
 M5-T &  0.001 &        0.01 &  0.807461 &  0.003538 \\
   M5 &  0.001 &        0.05 &  0.814005 &  0.011264 \\
   M5 &  0.001 &        0.01 &  0.821597 &  0.008436 \\
   M5 &  0.001 &           0 &  0.832461 &  0.003997 \\
 M5-T &      0 &        0.01 &  0.833508 &  0.005652 \\
   M5 &      0 &        0.05 &  0.835209 &  0.001013 \\
   M5 &      0 &         0.1 &  0.845812 &  0.001624 \\
   M5 &      0 &        0.01 &  0.866361 &  0.000427 \\
   M5 &      0 &           0 &  0.871597 &  0.000332 \\
\bottomrule
\end{tabular}
\end{center}
\end{table}

\end{document}